\documentclass[10pt,twocolumn,letterpaper]{article}

\usepackage[pagenumbers]{cvpr}

\usepackage[dvipsnames]{xcolor}

\usepackage{listings}
\definecolor{promptbg}{gray}{0.95}
\lstdefinestyle{promptstyle}{
    basicstyle=\ttfamily\scriptsize,
    breaklines=true,
    breakatwhitespace=false,
    breakindent=0pt,
    columns=fullflexible,
    keepspaces=true,
    frame=none,
    backgroundcolor=\color{promptbg},
    framexleftmargin=3pt,
    framexrightmargin=3pt,
    xleftmargin=3pt,
    xrightmargin=3pt,
    aboveskip=6pt,
    belowskip=6pt,
    captionpos=t,
    abovecaptionskip=4pt,
    belowcaptionskip=2pt,
    showstringspaces=false,
}

\newcommand{\blfootnote}[1]{%
    \begingroup
    \renewcommand{\thefootnote}{}%
    \renewcommand{\thempfootnote}{}%
    \footnotetext{#1}%
    \endgroup
}

\usepackage{colortbl} 
\definecolor{tabfirst}{rgb}{1, 0.7, 0.7} 
\definecolor{tabsecond}{rgb}{1, 0.85, 0.7} 
\definecolor{tabthird}{rgb}{1, 1, 0.7} 

\definecolor{cvprblue}{rgb}{0.21,0.49,0.74}
\usepackage[pagebackref,breaklinks,colorlinks,citecolor=cvprblue]{hyperref}

\usepackage[ruled,vlined,linesnumbered]{algorithm2e}
\usepackage{graphicx}
\usepackage{multirow}
\usepackage{booktabs}
\usepackage{threeparttable}

\def\paperID{245} 
\def\confName{3DV\xspace}
\def\confYear{2027\xspace}

\title{SceneMosaic: Efficient and Diverse Simulation-Ready Scene Generation via Hybrid Agentic Layout Evolution}

\author{
Xingjian Ran$^{1*}$ \quad
Xiaoye Mo$^{2*}$ \quad
Sihao Liu$^{1}$ \quad
Jianyu Zhang$^{1}$ \quad
Li Luo$^{1}$ \quad
Bo Dai$^{1\dagger}$ \\
$^{1}$The University of Hong Kong \quad
$^{2}$University of Electronic Science and Technology of China
}

\begin{document}
\twocolumn[{%
\renewcommand\twocolumn[1][]{#1}%
\maketitle
\vspace{-3.0em}
\begin{center}
\textbf{Project Page:} \url{https://rxjfighting.github.io/SceneMosaic}
\end{center}
\begin{center}
\captionsetup{type=figure}\includegraphics[width=\textwidth]{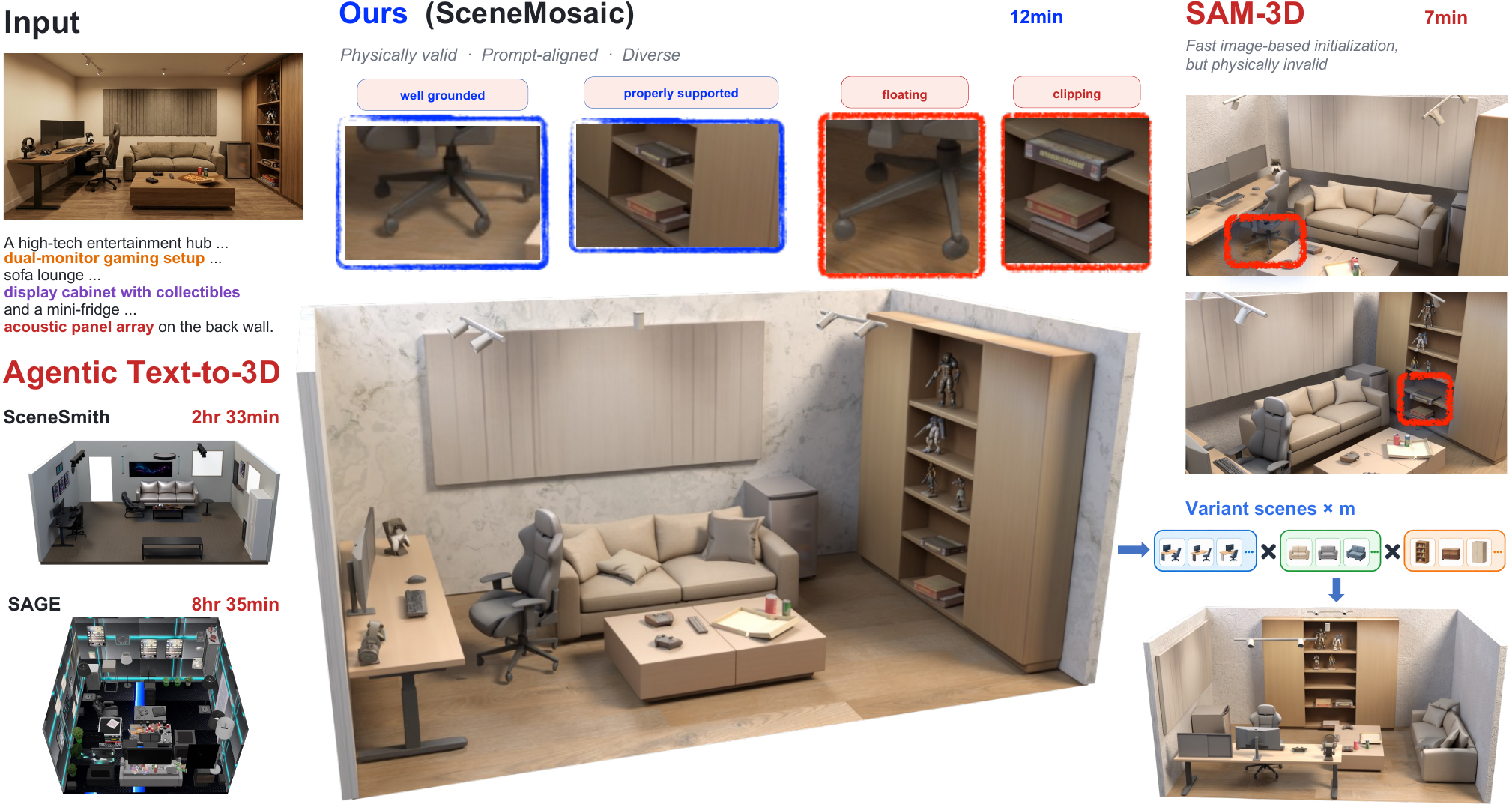}
\captionof{figure}{SceneMosaic efficiently generates simulation-ready 3D scenes, along with diverse layout variants, from a single reference image.}
\label{fig:teaser}
\end{center}%
}]
\blfootnote{$^{*}$Equal contribution.}
\blfootnote{$^{\dagger}$Corresponding author.}
\begin{abstract}

Diverse and simulation-ready indoor scenes are essential for interactive entertainment and embodied AI, yet their scalable generation remains challenging. 
Recent agentic text-to-3D scene pipelines that rely on vision-language models (VLMs) can generate scenes of high fidelity but require costly iterative object placement and refinement. 
Another mainstream paradigm, parametric image-to-3D scene models, produces scenes efficiently from strong priors learned from 2D images but often leads to imprecise and physically invalid scenes. 
More importantly, both paradigms struggle to output diverse scenes for a single input, making it hard for them to reflect the dynamically changing nature of real scenes.
In this paper we propose \textbf{SceneMosaic}, a framework that combines the merits of both paradigms.
It obtains the initial candidate from the learned image-based prior,
and subsequently evolves the result through VLM agents,
ensuring both efficiency and physical validity. 
Within the evolution process, \textbf{SceneMosaic} exploits the locality of natural scenes and decomposes a scene into independent local units,
allowing separate evolution within each unit before composing the global scene via Cartesian product.
On SceneEval-100, \textbf{SceneMosaic} matches the strongest agentic baseline in semantic layout quality with a $24\times$ speedup, substantially reduces physical violations, and receives the highest human ratings.
Our code is publicly available at \url{https://github.com/rxjfighting/SceneMosaic}.
\end{abstract}
\vspace{-2.0em}

\section{Introduction}

Scalable generation of simulation-ready indoor environments that faithfully reflect the density, clutter, and physical complexity of real ones is of great value in interactive entertainment and embodied AI. For example, general-purpose robots that can be deployed in arbitrary human houses must be trained and evaluated at a scale that real-world data collection cannot support, making simulation an indispensable substitute, yet existing simulation environments are sparsely furnished with limited diversity \cite{kolve2017ai2, li2023behavior, deitke2022, zhong2025internscenes}. 

Two major generative paradigms have emerged for simulation-ready indoor environments. \emph{Agentic text-to-3D scene pipelines}~\cite{fu2024anyhome, aguina2024open, yang2024holodeck, ccelen2024design, deng2025global, hsu2025programs} take a description and construct a scene through an agentic process, in which vision-language models (VLMs)~\cite{hurst2024gpt, qwen2.5-VL, google_gemini3_flash_2025} iteratively propose, place, and adjust objects while managing tool-use, memory, and context across the generation trajectory. As VLM capability has improved, such pipelines have become increasingly capable of producing high-quality, physically valid scenes despite the scarcity of large-scale 3D-native training data. Alternatively, \emph{parametric image-to-3D scene models} take a reference image as input, regarding scene construction as per-object asset generation and 6D pose estimation. This factorization is easier to supervise procedurally and benefits from the realistic visual layout priors present in images. Building on strong feed-forward asset generators~\cite{xiang2024structured, zhao2025hunyuan3d, lai2025hunyuan3d}, this has enabled a growing body of recent work~\cite{huang2025midi, meng2026scenegen, chen2026sam} that generates object assets and coarse scene layouts from images without iteration.

Despite the promising advances of existing paradigms, we identify two key intrinsic obstacles that limit their scalability and diversity.
The first limitation is a substantial efficiency–fidelity trade-off. Agentic text-to-3D pipelines achieve high-quality scenes, but those relying on iterative feedback loops require tool-intensive operations on each object to converge. Consequently, state-of-the-art methods like SceneSmith~\cite{pfaff2026scenesmith} and SAGE~\cite{xia2026sage} require multi-hour generation trajectories for a single scene, posing a key barrier to large-scale simulation.
Parametric image-to-3D scene models, by contrast, produce scenes efficiently via their learned strong layout priors, yet the generated object poses are often imprecise and physically invalid. 
The second limitation is the lack of diversity, as existing methods generate only a deterministic layout per input.
However, real-world scenes are dynamic and constantly reshaped by human activities, which reorient, relocate, and rearrange objects while preserving the scene's functional and physical plausibility.
Simulation benefits from multiple plausible layout variants capturing structured rearrangements, as exposing learners to such diversity broadens the environmental conditions covered during training and evaluation.
Although existing methods may obtain different layouts through repeated sampling, they rarely capture such structured rearrangements that preserve scene coherence.

To address these obstacles, in this paper we propose \textbf{SceneMosaic} (Figure~\ref{fig:teaser}), a hybrid framework that combines the complementary strengths of the two paradigms.
To bridge the efficiency–fidelity gap, \textbf{SceneMosaic} takes an image as input and leverages its image-based layout prior for rapid object pose initialization via a parametric image-to-3D scene model.
The coarse layout is then refined through agentic evolution to improve semantic and physical validity while retaining the efficiency.
Furthermore,
to capture the diversity of real scenes, in the agentic layout evolution process we exploit the locality of scene layouts and decompose them into independent local units, so that each local unit can be adjusted separately.
In this way, instead of producing a single candidate scene, \textbf{SceneMosaic} is capable of synthesizing a large set of candidate scenes through the Cartesian combination of local unit variants.
To enhance diversity while avoiding redundant candidates,
pairwise layout novelty is measured to greedily select novel candidates.
Finally, we conduct agentic layout evolution in orthographic 2D views, which remove perspective distortion and reduce pose editing to 2D translations and in-plane rotations, thereby simplifying spatial reasoning and promoting locally consistent edits.

Our key contributions are summarized as follows:
\begin{itemize}[left=1.5mm]
    \item We introduce \textbf{SceneMosaic}, a hybrid scene generation framework that combines image-based layout priors for fast scene initialization with agentic iterative evolution, yielding scenes that are simultaneously efficient to generate and physically and semantically valid.
    \item We propose an agentic layout evolution scheme that exploits the locality of scene layouts: each scene is decomposed into independent local units, whose layout variants are generated and refined separately and then composed via the Cartesian product, turning a single evolution pass into a combinatorial number of candidate scenes. A novelty-aware scene metric with dynamic representative selection is further devised to distill this large candidate pool into a compact yet diverse set of layouts.
    \item We show through extensive experiments on SceneEval-100 that SceneMosaic matches the strongest agentic baseline in semantic layout quality with a $24\times$ speedup, while substantially reducing the physical violations that plague image-to-3D models. A user study further confirms that our scenes are rated highest in both semantic and physical plausibility.
\end{itemize}
\section{Related Work}

\paragraph{3D Indoor Scene Synthesis.}
Early work learns object arrangements from annotated layout datasets~\cite{fu20213d}, using autoregressive transformers~\cite{paschalidou2021atiss, wang2021sceneformer}, diffusion models~\cite{tang2024diffuscene, zhai2024echoscene}, or explicit layout priors and constraint graphs~\cite{para2021generative, leimer2022layoutenhancer, lin2024instructscene} to place furniture within a room. A complementary line composes full 3D scenes with explicit geometry or radiance representations conditioned on text or scene graphs~\cite{zhang2024towards, gao2024graphdreamer, zhou2024gala3d, feng2025casagpt}, while procedural approaches synthesize environments through hand-crafted or learned rules~\cite{deitke2022, raistrick2024infinigen, li2025proc, ran2026pair2scene}, offering scale but limited semantic controllability. In contrast, our framework couples strong visual layout priors with agentic refinement, producing dense and physically valid scenes without category-specific training data.

\vspace{-0.7em}

\paragraph{LLM- and VLM-Driven Scene Generation.}
The strong reasoning and open-vocabulary capabilities of LLMs and VLMs~\cite{hurst2024gpt, qwen2.5, qwen2.5-VL, radford2021learning} have motivated language-guided scene generation, either by directly predicting numerical layouts or structured scene descriptions~\cite{feng2023layoutgpt, yang2024llplace, ran2025direct, sun2025hierarchically} or through program synthesis over object databases~\cite{aguina2024open, hsu2025programs}. More recent agentic pipelines orchestrate perception, placement, and iterative correction through tool use and multimodal feedback~\cite{yang2024holodeck, fu2024anyhome, ccelen2024design, sun2025layoutvlm, deng2025global, pun2025hsm, hao2025mesatask, yang2025sceneweaver, xia2026sage, pfaff2026scenesmith}, achieving strong semantic fidelity at substantial inference cost. Our approach retains the flexibility of agentic reasoning while grounding it in an image-based initialization and decomposing the scene into locally independent units for efficient and diverse evolution.

\vspace{-0.7em}

\paragraph{Image-to-3D Generation.}
Advances in feed-forward 3D asset generation~\cite{zhang2024clay, xiang2024structured, zhao2025hunyuan3d, lai2025hunyuan3d, zhang2026pixarmesh} have enabled recent image-to-3D scene methods that jointly infer per-object geometry and coarse layouts from a reference image~\cite{huang2025midi, meng2026scenegen, chen2026sam}, benefiting from realistic visual priors and fast initialization. However, the resulting object poses are frequently imprecise, leading to collisions and boundary violations. We leverage these priors for rapid scene initialization and refine object poses through agentic layout evolution to enforce physical plausibility.
\section{Method}

\begin{figure*}[t] \centering \includegraphics[width=\linewidth]{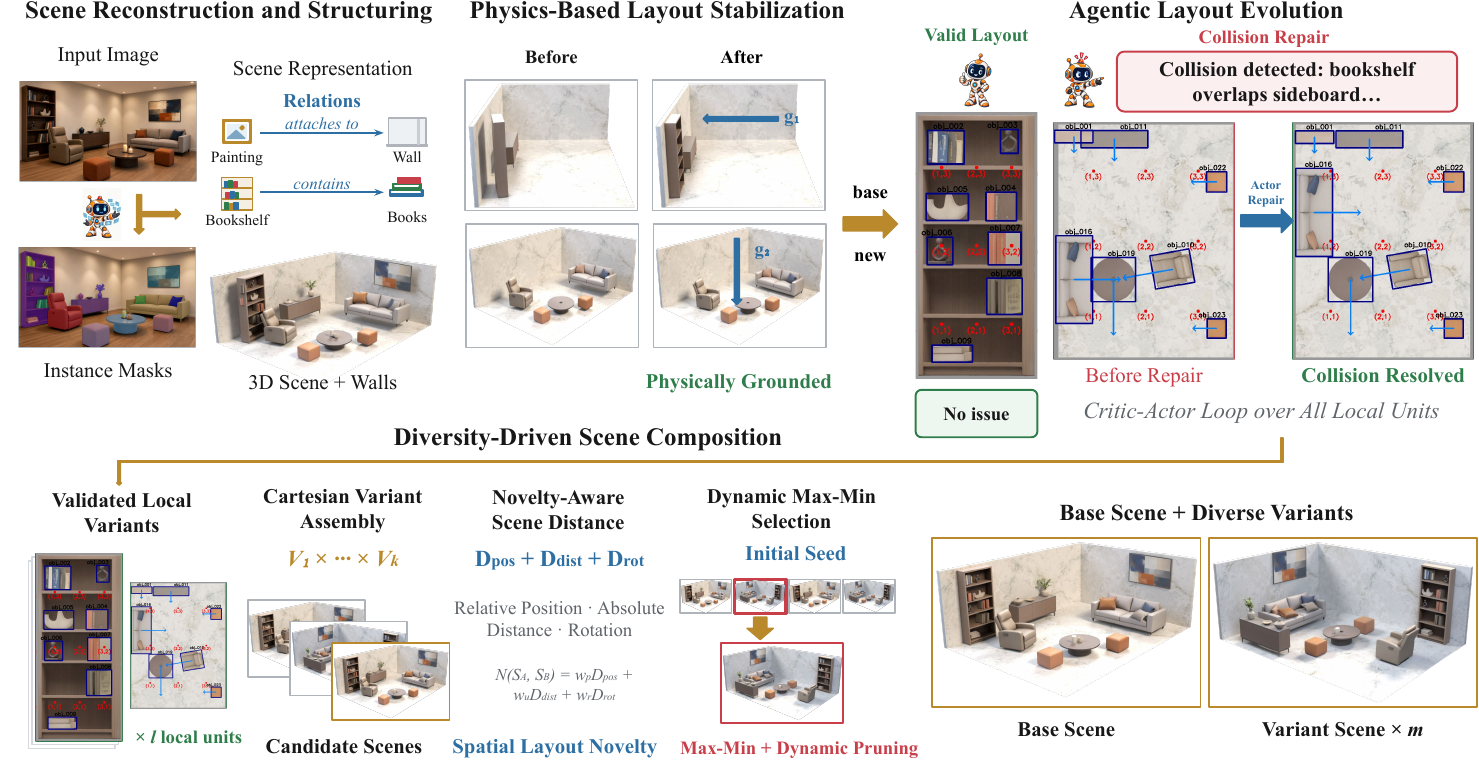} \caption{Overview of our proposed framework. Given a single reference input signal, we first reconstruct and structure the scene into object-centric local units. We then perform physics-based layout stabilization and Critic-Actor agentic evolution to generate diverse local layout variants. Finally, the variants are combinatorially assembled and selected using a novelty-aware distance metric and dynamic greedy search to produce diverse, physically plausible scene layouts.}
\label{fig:pipeline}
\vspace{-1.0em}
\end{figure*}

\subsection{Problem Formulation}
Given a single reference input signal (e.g., an image or a text-to-image prompt), our goal is to generate a simulation-ready base scene and $m$ diverse variants. Figure~\ref{fig:pipeline} provides an overview of the proposed framework.

Formally, a 3D scene $S$ is composed of a set of $N$ objects $\mathcal{O} = \{o_1, o_2, \dots, o_N\}$. Each object $o_i$ is parameterized by its 3D mesh representation $\mathcal{M}_i$ alongside its 3D spatial layout parameters $\mathbf{T}_i = (\mathbf{p}_i, \mathbf{R}_i, \mathbf{s}_i)$, and additional simulation-related properties, where $\mathbf{p}_i \in \mathbb{R}^3$ denotes the 3D translation (position), $\mathbf{R}_i \in \mathrm{SO}(3)$ represents the rotation matrix (or equivalent unit quaternion $\mathbf{q}_i \in \mathbb{S}^3$)~\cite{zhou2019continuity}, and $\mathbf{s}_i \in \mathbb{R}^3$ defines the anisotropic scale.

\subsection{Scene Reconstruction and Structuring}
\paragraph{Object-Centric Scene Reconstruction.}
We first unify all input signals into a scene image. Based on the image, we construct an initialized 3D scene that serves as the basis for subsequent relation reasoning and layout evolution. Specifically, a perception agent first performs instance-level object registration, producing a stable object inventory with semantic labels, textual descriptions, and 2D bounding boxes. Each registered object is then segmented by SAM3~\cite{carion2026sam} using its textual prompt and bounding box. To improve segmentation quality, the perception agent performs an iterative refinement that resolves missing instances, duplicated masks, ambiguous boundaries, and nested object regions through iterative mask verification and re-segmentation. After refinement, each object is represented by a verified instance mask while maintaining a unified object identity.

Finally, SAM3D~\cite{chen2026sam} reconstructs each object independently from the scene image and its verified mask, producing an object mesh $\mathcal{M}_i$ together with its initialized layout parameters $\mathbf{T}_i=(\mathbf{p}_i,\mathbf{R}_i,\mathbf{s}_i)$. The resulting initialized scene therefore provides a consistent correspondence between semantic object identities, pixel-level evidence, reconstructed meshes, and spatial layouts.

\vspace{-0.7em}

\paragraph{Relation-Guided Scene Structuring.}
The reconstructed objects are further converted into a structured scene representation by jointly inferring room structures, object relations, and physical properties. Given the initialized layouts, object masks, and the multi-view renderings of the scene, we recover a canonical room boundary, instantiate wall primitives, and identify object relations including \textit{attach} (e.g., floor, wall, ceiling, or object support) and \textit{contain} dependencies between objects.

To accommodate heterogeneous reasoning tasks, we organize the relation extraction process as a directed acyclic graph (DAG). In this framework, a manager agent first analyzes the scene and dynamically schedules specialized task agents for support reasoning, containment analysis, wall recovery, wall attachment, semantic refinement, and other related subtasks. Each subagent operates on task-specific visual and geometric evidence, while DAG dependencies guarantee consistent information flow between tasks. In parallel, simulation attributes are estimated for each object.

Based on the inferred attach and contain dependencies, and exploiting the modular structure of indoor layouts with locally independent regions, the scene is structured as a hierarchical scene tree. We explicitly define each \textbf{local unit} as a sub-scene composed of a non-leaf node (serving as an anchor) together with all of its immediate child nodes. Local units can be evolved largely independently, but a small number of functional relations span unit boundaries. We therefore extract cross-unit constraints and integrate them as unit-local guidance into the corresponding contexts. For example, floor-standing chairs are constrained to face the desk, while desk-mounted monitors preserve the desk orientation. Finally, a verification pass resolves inconsistencies and validates completeness, establishing a grounded scene with well-defined local units to guide layout evolution.

\subsection{Agentic Layout Evolution}

\paragraph{Physics-Based Layout Stabilization.}
Before agentic evolution, we further optimize the physical consistency of each local layout following the top-down hierarchy of the scene tree. For each local unit, we first perform containment correction by checking whether each target object is enclosed by its anchor object in the orthographic projection and adjusting invalid layouts accordingly to establish a solid physical foundation. We then conduct gravity-based simulation by placing the local unit into a physical simulation environment with an appropriate gravity direction, enabling objects to resolve collisions and eliminate unsupported floating states. This process produces physically grounded local layouts that provide a stable initialization for subsequent agentic evolution.

\vspace{-0.7em}

\paragraph{Critic-Actor Layout Evolution.}
Starting from the physically stable local layouts obtained earlier, we further improve their quality while capturing the diversity of scenes through an agentic evolution process. Specifically, for each local unit, we first spawn a set of diverse candidate variants from its original layout, after which both the original layout and the generated variants undergo parallel agentic evolution. Within each evolution step, we employ a Critic-Actor agentic loop: The Critic utilizes visual evidence and simulation tools to assess layout validity, returning diagnostic feedback and targeted optimization suggestions. The Actor updates the local spatial layout parameters based on the Critic's suggestions.
Upon convergence, the refined local layouts and their variants undergo a final simulation to guarantee strict physical plausibility.

\vspace{-0.7em}

\paragraph{Agent Memory and Spatial Abstraction.}
To stabilize spatial reasoning and prevent agents from falling into optimization deadlocks, we refine the context and memory management mechanisms. First, the context of each agent is strictly scoped to local spatial information relevant only to its target local unit. Second, rather than rendering perspective 3D views, we project each local layout into an appropriate orthographic 2D representation, reducing 3D pose adjustments to 2D center translations and in-plane rotations. Third, we maintain an explicit history summary of past suggestions and modifications, enabling agents to detect cyclic adjustments and escape local loops.

\subsection{Diversity-Driven Scene Composition}

\paragraph{Combinatorial Variant Assembly.}
Following local evolution, every local unit yields a set of validated, physically sound local layout variants. We then perform a Cartesian combination across all local unit variants, producing $k$ globally complete candidate scenes. To mitigate scene homogenization and manage the exponential candidate pool, we evaluate pairwise scene dissimilarity using a quantitative layout novelty metric and select a final representative subset of $m+1$ diverse scenes using a dynamic greedy search strategy.

\vspace{-0.7em}

\paragraph{Novelty-Aware Scene Distance.}
Let $S_A$ and $S_B$ be two scenes composed of the same set of common objects $\mathcal{O} = \{o_1, o_2, \dots, o_N\}$. For each object $o_i \in \mathcal{O}$, let $\mathbf{p}_i \in \mathbb{R}^3$, $\mathbf{R}_i \in \mathrm{SO}(3)$, and $\mathbf{q}_i \in \mathbb{S}^3$ denote its position, rotation matrix, and unit quaternion orientation, respectively. We define the novelty-aware scene distance by combining three normalized difference measures:

\textbf{Relative Position Difference ($D_{\mathrm{pos}}$):}
To capture relative spatial arrangement invariant to global rigid transformations, pairwise displacement vectors between objects $i$ and $j$ ($i \neq j$) are projected into object $i$'s local coordinate frame:
\begin{equation}
    \mathbf{v}_{ij, \text{local}}^{(A)} = (\mathbf{R}_i^{(A)})^T (\mathbf{p}_j^{(A)} - \mathbf{p}_i^{(A)}).
\end{equation}
The normalized relative position difference is defined as:
\begin{equation}
    D_{\mathrm{pos}} = \biggr( \frac{\sum_{i \neq j} \|\mathbf{v}_{ij,\text{local}}^{(A)} - \mathbf{v}_{ij,\text{local}}^{(B)}\|_2^2}{\sum_{i \neq j} \|\mathbf{p}_j^{(A)} - \mathbf{p}_i^{(A)}\|_2^2} \biggr)^{\!1/2}.
\end{equation}

\textbf{Absolute Distance Difference ($D_{\mathrm{dist}}$):}
To quantify changes in global inter-object proximity, let
$
d_{ij}^{(X)}=\|\mathbf{p}_j^{(X)}-\mathbf{p}_i^{(X)}\|_2,
$
denote the pairwise Euclidean distances between objects in scene $X$. We define
\begin{equation}
D_{\mathrm{dist}}
=
\biggr(
\frac{
\sum_{i\neq j}
(d_{ij}^{(A)}-d_{ij}^{(B)})^2
}{
\sum_{i\neq j}
(d_{ij}^{(A)})^2
}
\biggr)^{1/2}.
\end{equation}

\textbf{Rotation Difference ($D_{\mathrm{rot}}$):}
The orientation difference per object is measured via the quaternion inner product $\theta_i = 2 \arccos(|\mathbf{q}_i^{(A)} \cdot \mathbf{q}_i^{(B)}|) \in [0, \pi]$. The normalized rotation difference is:
\begin{equation}
    D_{\mathrm{rot}} = \frac{1}{\pi N} \sum_{i=1}^N \theta_i.
\end{equation}

The overall scene distance $\mathcal{N}(S_A, S_B)$ between the two scenes is defined as:
\begin{equation}
    \mathcal{N}(S_A, S_B) = w_p D_{\mathrm{pos}} + w_u D_{\mathrm{dist}} + w_r D_{\mathrm{rot}},
\end{equation}
where $w_p + w_u + w_r = 1$ and $S_A$ is the reference scene.

\vspace{-0.7em}

\paragraph{Greedy Diverse Selection.}
To select $m$ highly diverse yet realistic variant scenes from the candidate pool $\mathcal{C}$, we employ a Dynamic Max-Min Greedy Search, detailed in the supplementary material. The selection process operates under three key principles: the base scene $S_{\text{org}}$ is first retained as the initial seed to guarantee the inclusion of the reference layout. Subsequently, at each iteration, the candidate that maximizes the minimum novelty-aware distance to the already selected set $\mathcal{R}$ is added, thereby encouraging maximal spatial diversity among the selected scenes. After each new selection, candidates whose minimum distance to the selected set falls below a dynamic threshold $\tau=\alpha \cdot d_{\max}$ with respect to the current selected set are pruned, preventing localized clustering while accelerating the convergence of the search.
\section{Experiments}

\subsection{Experimental Setup}

\begin{figure*}[t]
    \centering
    \includegraphics[width=\linewidth]{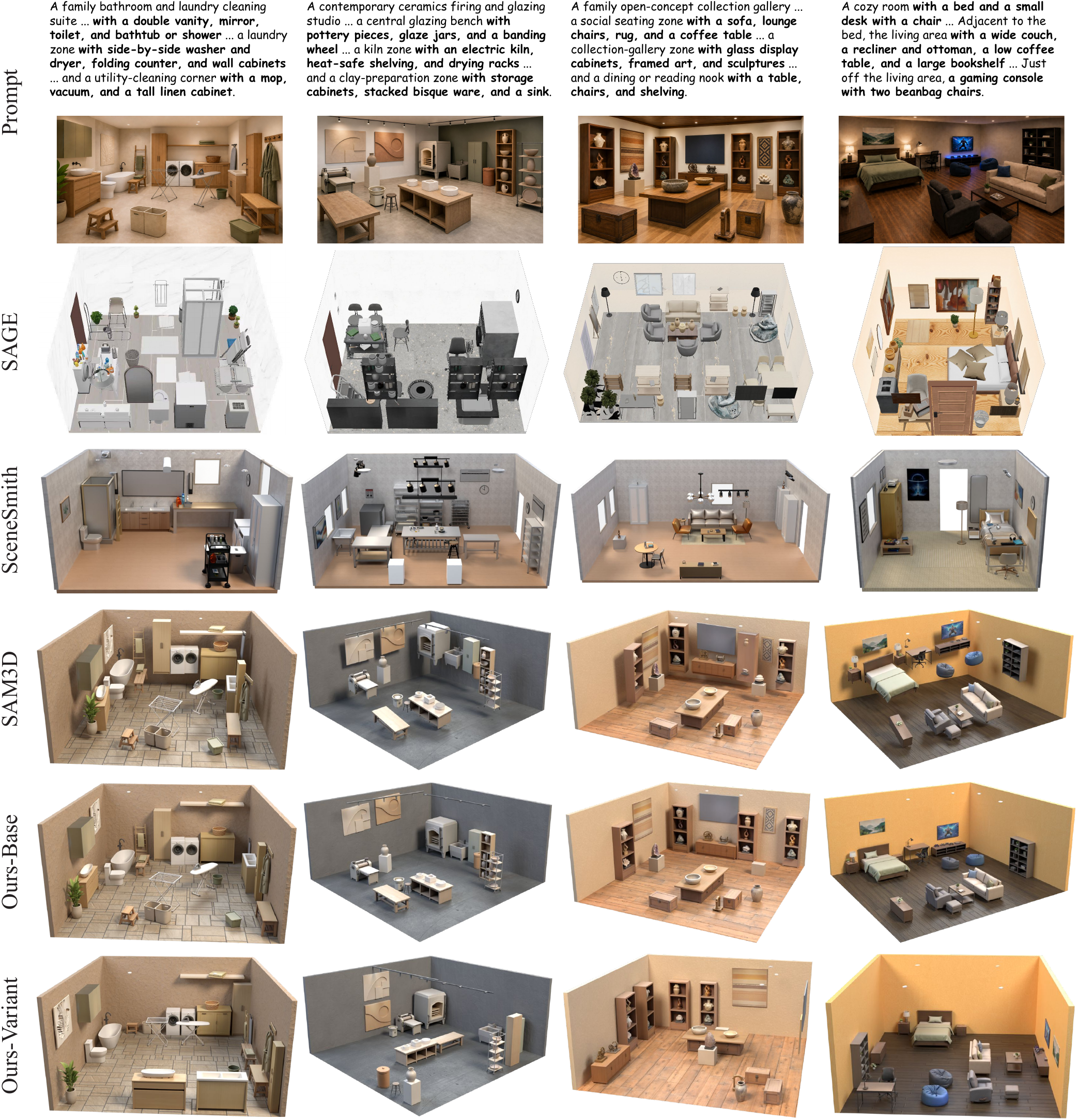}
    \caption{Qualitative comparison with existing text-to-3D and image-to-3D baseline methods. Our approach generates realistic, high-quality 3D layouts and supports fast variant scene generation while avoiding physically implausible configurations.}
    \label{fig:qualitative}
\vspace{-1.0em}
\end{figure*}

\begin{table*}[t]
\centering
\begin{threeparttable}
\caption{Quantitative comparison on SceneEval-100. POS and ROT evaluate semantic layout quality, while NAV, COL, and OOB measure physical validity. Time indicates the average generation time per scene.}
\label{tab:quantitative}
\begin{tabular}{l|ccccc|c}
\toprule
Method & POS $\uparrow$ & ROT $\uparrow$ & NAV $\uparrow$ & COL $\downarrow$ & OOB $\downarrow$ & Time $\downarrow$ \\
\midrule
\multicolumn{7}{l}{\textit{Agentic Text-to-3D}} \\
HoloDeck & 49.8$\pm$6.2 & 47.2$\pm$7.6 & \cellcolor{tabfirst}99.5$\pm$0.3 & 12.1$\pm$3.4 & 0.9$\pm$0.3 & \cellcolor{tabsecond}0.05h \\
SceneWeaver & 55.2$\pm$6.8 & 51.8$\pm$6.5 & 97.3$\pm$1.1 & 10.7$\pm$3.2 & \cellcolor{tabfirst}0.0$\pm$0.0 & 1.07h \\
SAGE\tnote{$\dagger$} & 59.8$\pm$7.9 & 53.3$\pm$5.5 & 90.1$\pm$1.2 & \cellcolor{tabthird}1.8$\pm$2.5 & \cellcolor{tabthird}0.8$\pm$0.5 & 7.56h \\
SceneSmith & \cellcolor{tabthird}83.5$\pm$2.7 & \cellcolor{tabfirst}80.9$\pm$3.1 & 96.9$\pm$0.8 & \cellcolor{tabsecond}0.7$\pm$1.4 & \cellcolor{tabsecond}0.3$\pm$0.3 & 3.43h \\
\midrule
\multicolumn{7}{l}{\textit{Image-to-3D}} \\
MIDI & 76.3$\pm$5.2 & 71.8$\pm$6.4 & 98.4$\pm$1.1 & 19.8$\pm$5.7 & N/A\tnote{$\ddagger$} & 0.19h \\
SceneGen & 74.7$\pm$5.8 & 70.5$\pm$7.2 & 95.1$\pm$1.4 & 25.6$\pm$6.9 & N/A\tnote{$\ddagger$} & 0.19h \\
SAM3D & 80.1$\pm$4.5 & 75.4$\pm$7.0 & 98.9$\pm$0.7 & 20.2$\pm$6.1 & 8.9$\pm$4.2 & \cellcolor{tabsecond}0.05h \\
\midrule
\multicolumn{7}{l}{\textit{Ours}} \\
Base Scenes & \cellcolor{tabsecond}84.3$\pm$2.0 & \cellcolor{tabthird}78.4$\pm$3.4 & \cellcolor{tabthird}99.2$\pm$0.4 & \cellcolor{tabfirst}0.0$\pm$0.0 & \cellcolor{tabfirst}0.0$\pm$0.0 & \cellcolor{tabthird}0.14h \\
Variant Scenes & \cellcolor{tabfirst}84.5$\pm$2.4 & \cellcolor{tabsecond}79.9$\pm$3.6 & \cellcolor{tabsecond}99.4$\pm$0.3 & \cellcolor{tabfirst}0.0$\pm$0.0 & \cellcolor{tabfirst}0.0$\pm$0.0 & \cellcolor{tabfirst}0.03h \\
\bottomrule
\end{tabular}

\begin{tablenotes}
\footnotesize
\item[$\dagger$] SAGE runs locally deployed models rather than external APIs.
\item[$\ddagger$] N/A: MIDI and SceneGen do not recover room boundaries, so the out-of-bounds rate is not applicable.
\end{tablenotes}

\end{threeparttable}
\vspace{-0.5em}
\end{table*}

\paragraph{Baselines.}
We compare our approach with two representative paradigms of scene generation methods. The first paradigm consists of agentic text-to-3D scene generation methods, including HoloDeck~\cite{yang2024holodeck}, SceneWeaver~\cite{yang2025sceneweaver}, SAGE~\cite{xia2026sage}, and SceneSmith~\cite{pfaff2026scenesmith}. These approaches mainly rely on vision-language models and iterative reasoning pipelines to construct scenes from textual descriptions. The second paradigm consists of image-to-3D generation methods, including MIDI~\cite{huang2025midi}, SceneGen~\cite{meng2026scenegen}, and SAM3D~\cite{chen2026sam}, which leverage visual input and learned priors to directly infer object assets and layouts. Note that while baseline methods are evaluated solely on single-scene generation, we evaluate our approach on both the generated base scenes (i.e., scenes corresponding to the input) and their variants to demonstrate generation quality and variation capabilities. For variant scenes, we consistently set $m = 5$, generating 5 variant scenes for each base scene. 

\vspace{-0.7em}

\paragraph{Evaluation Protocol.}
We evaluate our method using 100 manually designed indoor scene prompts from SceneEval-100~\cite{tam2026sceneeval} for a unified comparison. Since our method takes image inputs while the final objective is simulation-ready scene generation rather than scene reconstruction, we compare different approaches across input modalities using modality-independent generation metrics. Specifically, we use the provided text prompts as inputs for text-to-3D baselines, while the corresponding generated images are used as inputs for image-to-3D baselines and our method. This protocol aims to ensure as fair an evaluation as possible across different scene generation paradigms under the same generation objectives. All experiments and baseline comparisons are conducted in the same software and hardware environment using a single NVIDIA L40S GPU. The only exception is SAGE, which utilizes 5 NVIDIA H200 GPUs to accommodate the local deployment of its VLM agents.

\vspace{-0.7em}

\paragraph{Metrics.}
We assess generated scenes from both semantic and physical perspectives. Following prior work~\cite{sun2025layoutvlm}, we leverage GPT-5.5 to check if objects are placed with semantically reasonable positions (POS) and rotations (ROT). In addition, we adopt three physical validity metrics from SceneEval-100~\cite{tam2026sceneeval}: navigability (NAV), collision rate (COL), and out-of-bounds rate (OOB). Navigability measures whether generated layouts preserve feasible navigation spaces, while collision rate and out-of-bounds rate evaluate physical violations caused by object collisions and objects being out of bounds.

\subsection{Quantitative Results}

Table~\ref{tab:quantitative} reveals clear trade-offs among efficiency, semantic quality, and physical validity in existing 3D scene generation paradigms. Although agentic text-to-3D methods (e.g., SceneWeaver, SAGE) can construct complex scenes via iterative reasoning and tool calling, they suffer from limited instruction-following and high latency. Although SceneSmith achieves high performance metrics, its prolonged generation time can be even slower than manual 3D editing. HoloDeck, which operates without feedback loops, accelerates generation via layout search, yet lacks global semantic guidance. Conversely, image-to-3D models (e.g., MIDI, SceneGen, SAM3D) achieve rapid initialization, but direct prediction leads to severe physical violations.

Our method bridges these paradigms by combining fast image-based initialization with agentic layout evolution for spatial and physical refinement. Consequently, it achieves competitive semantic quality, significantly alleviates physical violations, and drastically reduces inference overhead compared to agentic baselines while outperforming image-to-3D models in quality and variant generation speed.

\subsection{Qualitative Results}

Figure~\ref{fig:qualitative} presents qualitative comparisons between our method and recent state-of-the-art text-to-3D and image-to-3D scene generation methods. SAGE and SceneSmith are capable of producing semantically meaningful scenes through iterative reasoning, but the long optimization trajectory often results in expensive generation and may deviate from the original semantics. SAM3D provides efficient initialization by exploiting visual priors. However, the predicted object poses frequently contain physically implausible configurations, such as floating objects or interpenetrating structures.

In contrast, our method first inherits the global spatial structure from the input image and then performs agentic layout evolution to progressively improve object relations. This combination allows the generated scenes to maintain realistic semantic arrangements while satisfying physical constraints. Moreover, the proposed evolution strategy enables diverse variant scenes while preserving the coherence of the base scene, demonstrating its capability for scalable simulation-ready scene generation.

\begin{table}[t]
\centering
\caption{User study results. We report the mean and standard deviation on semantic plausibility (Sem.) and physical plausibility (Phy.), along with Kendall's coefficient.}
\label{tab:userstudy}
\small
\begin{tabular}{l|cc}
\toprule
Method & Sem. $\uparrow$ & Phy. $\uparrow$ \\
\midrule
SceneSmith & \cellcolor{tabsecond}4.01$\pm$0.99 & \cellcolor{tabsecond}4.16$\pm$0.97 \\
SAGE & 3.08$\pm$1.13 & \cellcolor{tabthird}2.81$\pm$1.31 \\
SceneGen & 2.12$\pm$1.21 & 2.08$\pm$1.33 \\
SAM3D & \cellcolor{tabthird}3.22$\pm$1.21 & 2.43$\pm$1.45 \\
Ours & \cellcolor{tabfirst}4.33$\pm$0.89 & \cellcolor{tabfirst}4.47$\pm$0.82 \\
\midrule
Kendall's $W$ & 0.648 & 0.633 \\
\bottomrule
\end{tabular}
\vspace{-1.5em}
\end{table}

\begin{table*}[t]
\centering
\caption{Ablation study on SceneEval-100. We evaluate the impact of key component choices on both base and variant scene generation.}
\label{tab:ablation}
\small
\begin{tabular}{l|ccccc|c}
\toprule
 & POS $\uparrow$ & ROT $\uparrow$ & NAV $\uparrow$ & COL $\downarrow$ & OOB $\downarrow$ & Time $\downarrow$ \\
\midrule
\multicolumn{7}{l}{\textit{\textbf{Base Scenes}}} \\
Full SceneMosaic & 84.3$\pm$2.0 & 78.4$\pm$3.4 & 99.2$\pm$0.4 & 0.0$\pm$0.0 & 0.0$\pm$0.0 & 0.14h \\
(1) w/o Image Init (Text-only Init) & 80.7$\pm$3.2 & 74.2$\pm$3.5 & 98.8$\pm$0.6 & 0.0$\pm$0.0 & 0.0$\pm$0.0 & 1.25h \\
(2) w/o Physics & 81.5$\pm$2.4 & 75.6$\pm$3.7 & 96.8$\pm$1.6 & 18.9$\pm$4.8 & 8.5$\pm$3.3 & 0.13h \\
(3) Perspective 3D Ops + Local Decomposition & 78.1$\pm$4.1 & 71.8$\pm$4.4 & 98.7$\pm$0.7 & 2.6$\pm$1.3 & 2.4$\pm$1.1 & 0.31h \\
(4) Global Evolution + Perspective 3D Ops & 72.3$\pm$4.8 & 64.9$\pm$5.6 & 97.2$\pm$1.3 & 12.1$\pm$3.7 & 6.2$\pm$2.0 & 0.68h \\
\midrule
\multicolumn{7}{l}{\textit{\textbf{Variant Scenes}}} \\
Full SceneMosaic & 84.5$\pm$2.4 & 79.9$\pm$3.6 & 99.4$\pm$0.3 & 0.0$\pm$0.0 & 0.0$\pm$0.0 & 0.03h \\
(1) w/o Image Init (Text-only Init) & 83.9$\pm$2.9 & 77.6$\pm$4.0 & 99.3$\pm$0.4 & 0.0$\pm$0.0 & 0.0$\pm$0.0 & 0.26h \\
(2) w/o Physics & 81.3$\pm$2.2 & 74.8$\pm$3.2 & 97.1$\pm$1.4 & 18.7$\pm$4.5 & 8.3$\pm$3.2 & 0.03h \\
(3) Perspective 3D Ops + Local Decomposition & 78.2$\pm$3.7 & 71.5$\pm$3.9 & 98.2$\pm$0.6 & 1.3$\pm$0.5 & 0.6$\pm$0.2 & 0.06h \\
(4) Global Evolution + Perspective 3D Ops & 72.6$\pm$4.4 & 65.1$\pm$5.2 & 97.7$\pm$1.1 & 7.1$\pm$2.3 & 4.2$\pm$1.7 & 0.65h \\
\bottomrule
\end{tabular}
\vspace{-0.5em}
\end{table*}

\subsection{User Study}

We further conduct a user study comparing our method against the most recent baselines from both paradigms, where 48 participants rate 15 scenes (3 scenes per method $\times$ 5 methods) on semantic and physical plausibility using a 5-point Likert scale. As shown in Table~\ref{tab:userstudy}, our method achieves the highest scores on both dimensions with the lowest standard deviations. Human judgments align well with the automated metrics: SceneSmith ranks second at a much higher generation cost, while the direct layout predictions of SceneGen and SAM3D receive low physical plausibility scores due to frequent interpenetration. Kendall's coefficient of concordance indicates strong inter-rater agreement on both dimensions ($W \geq 0.633$, $p < 0.001$), confirming the reliability of these preferences.

\subsection{Ablation Study}

To validate the key designs in SceneMosaic, we conduct comprehensive ablation experiments on both base and variant scene generation using the SceneEval-100 benchmark. Specifically, as shown in Table~\ref{tab:ablation}, we investigate four architectural and operational variants. As no baseline generates multiple layouts from a fixed asset set, layout diversity is ablated in the supplementary material.

\textbf{w/o Image Init (Text-only Init):} Replacing image-based initialization with object-wise text-driven initialization while keeping the rest of the pipeline unchanged.

\textbf{w/o Physics:} Omitting both the physics-based pre-stabilization step (containment correction and gravity simulation) and all physical simulation tools utilized during agentic layout evolution.

\textbf{Perspective 3D Operations + Local Decomposition:} Replacing orthographic 2D spatial projections with perspective 3D views and 3D edit operations during agent reasoning, while maintaining local unit decomposition.

\textbf{Global Evolution + Perspective 3D Operations:} Performing agentic evolution globally across the entire scene under perspective 3D representations without local unit decomposition (note that global evolution is incompatible with 2D orthographic projection).

\vspace{-0.7em}

\paragraph{Impact of Image-Based Initialization.}
Removing image priors leads to a drastic increase in generation latency due to the extensive iterative search required by agentic text-to-3D processes to establish spatial layouts. Furthermore, text-based initialization causes slight degradation in semantic alignment scores, proving that grounding scene construction in realistic visual priors provides a stronger starting point for original layout refinement.

\vspace{-0.7em}

\paragraph{Role of Physics-Based Processing and Tools.}
Omitting both the physics pre-stabilization step and the simulation tools during layout evolution results in a noticeable degradation in physical validity metrics, leading to substantial increases in collision and out-of-bounds rates. While generation time remains relatively low, the lack of physical grounding allows object interpenetrations and invalid floating configurations to persist throughout the evolution process, demonstrating the indispensable role of physical tools and stabilization.

\vspace{-0.7em}

\paragraph{Orthographic 2D Abstraction vs. Perspective 3D Operations.}
Replacing 2D orthographic projections with 3D perspective representations negatively affects both efficiency and layout accuracy. Perspective views introduce visual occlusions and complex 3D transformation spaces, causing agents to struggle with spatial reasoning. This noticeably increases inference time while degrading semantic orientation and position quality.

\vspace{-0.7em}

\paragraph{Effectiveness of Local Unit Decomposition.}
Performing global evolution without local unit decomposition yields the lowest overall scores across semantic metrics. Reasoning over the entire scene simultaneously creates excessive context complexity, leading to elevated collision rates and slow convergence. Crucially, as shown in the lower part of Table~\ref{tab:ablation}, global evolution completely fails to accelerate variant scene generation, whereas our local decomposition framework enables combinatorial Cartesian assembly that generates diverse and realistic variants at a fraction of the cost.
\section{Conclusion}

In this paper, we presented SceneMosaic, a hybrid framework for efficient and diverse simulation-ready scene generation. It initializes scenes from image-based layout priors and refines them through agentic layout evolution over orthographic 2D projections, while the local unit decomposition enables Cartesian composition of diverse layout variants at negligible extra cost. Experiments on SceneEval-100 and a user study demonstrate that SceneMosaic matches or surpasses state-of-the-art baselines in semantic quality, eliminates physical violations such as collisions and out-of-bounds placements, and reduces generation time by an order of magnitude compared to state-of-the-art feedback-driven agentic pipelines.

{
    \small
    \bibliographystyle{ieeenat_fullname}
    \bibliography{main}
}

\clearpage
\setcounter{page}{1}
\maketitlesupplementary

\begin{figure*}[!t]
    \centering
    \includegraphics[width=\textwidth]{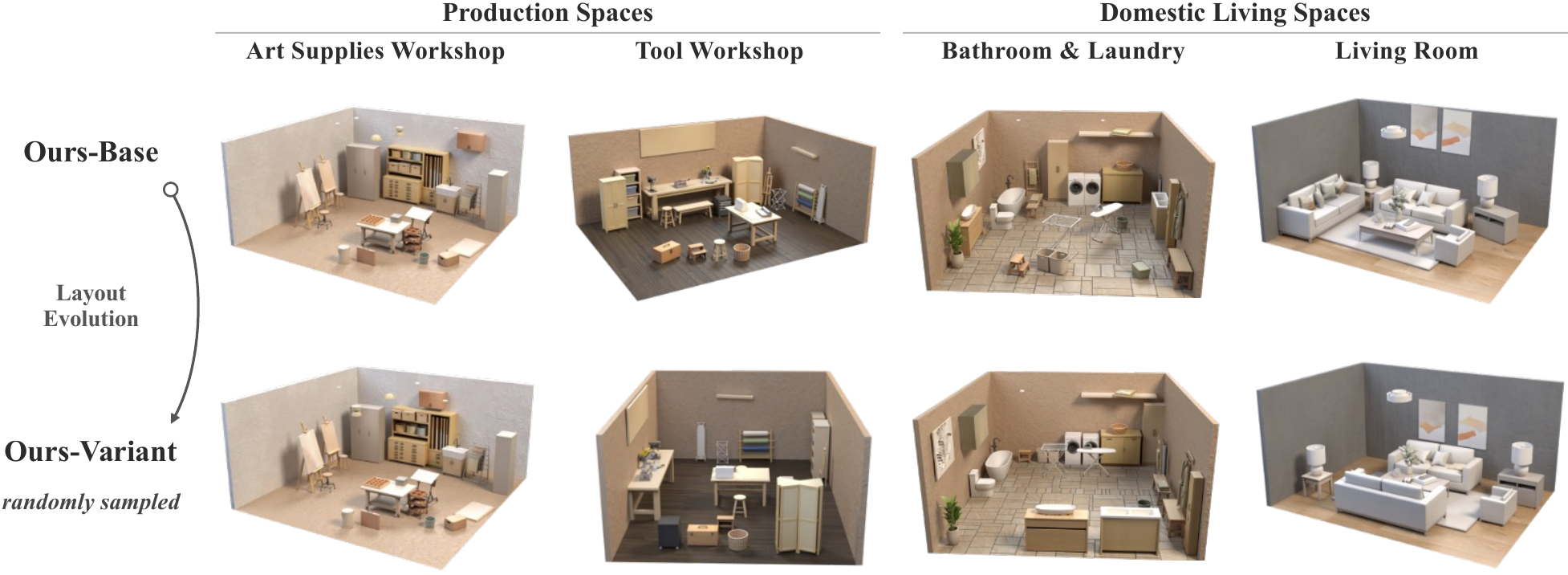}
    \caption{Additional qualitative results of our method. We randomly sample variant scenes generated by our framework to demonstrate the diversity of the resulting layouts. The examples show that our method can produce diverse yet physically plausible scene configurations while preserving the semantic structure of the input scene.}
    \label{fig:additional_results}
\end{figure*}

\section{Implementation Details}

Figure~\ref{fig:additional_results} presents additional qualitative results, showcasing randomly sampled variant scenes generated by our framework.

\paragraph{Agent Configuration.}
All agents in our framework, including the perception agent, the relation-reasoning task agents, and the Critic-Actor evolution agents, are instantiated from the same underlying VLM through role-specific prompts. Given a textual input, the reference scene image is synthesized by the text-to-image model GPT-Image-2; the images used in our experiments are generated from the SceneEval-100 prompts in this way. Each agent receives only the visual and structured evidence relevant to its target local unit, and all agent responses are constrained to structured JSON schemas that are parsed and validated before being applied to the scene. Since local units are mutually independent, both the evolution of different local units and the evolution of different variants within a unit are executed in parallel, which substantially reduces wall-clock generation time.

\vspace{-0.7em}

\paragraph{Physics Simulation.}
Physics-based stabilization and validation are performed in a rigid-body simulator. For each object $o_i$, we precompute a convex decomposition of its mesh $\mathcal{M}_i$ via CoACD to obtain watertight collision hulls, which are reused across all stabilization, evolution, and validation stages. The gravity direction is anchor-dependent: floor-anchored units use downward gravity $(0,-g,0)$, ceiling-anchored units use inverted gravity $(0,+g,0)$, and wall-anchored units use gravity aligned with the outward wall normal, so that attached objects settle naturally onto their supporting surface. During simulation, the anchor object and the room boundary primitives are treated as static ground bodies, while target objects are simulated as dynamic rigid bodies until their poses converge. Objects whose vertical extent is negligible relative to their planar extent (e.g., rugs and paper sheets) are excluded from collision counting to avoid spurious contact reports.

\vspace{-0.7em}

\paragraph{Orthographic Projection and Coordinate Mapping.}
For each local unit, we render a $1024 \times 1024$ orthographic view along the anchor-dependent projection direction (top-down for floor-anchored units, along the outward normal for wall-anchored units, and along the anchor's front direction for containment units). The rendering is augmented with (i) a uniform coordinate grid overlaid on the anchor surface, (ii) per-object 2D bounding boxes, and (iii) arrows indicating each object's facing direction. Every grid point stores its corresponding 3D location on the anchor surface, establishing an invertible mapping between normalized 2D coordinates $(u, v)$ and 3D positions. A 2D pose predicted by an agent, consisting of a center $(u, v)$ and an in-plane rotation angle, is thus deterministically lifted back to the 3D layout parameters $\mathbf{T}_i$: the center is mapped through the grid correspondence (with bilinear interpolation between grid points), and the in-plane angle is converted to a rotation about the projection axis composed with the object's original orientation. The anchor object itself is never modified during evolution.

\section{Details of Critic-Actor Layout Evolution}

\paragraph{Critic.}
At each evolution round, the Critic receives the current orthographic rendering, the structured 2D layout state (object centers, rotations, and bounding boxes in grid coordinates), the deep-collision pairs reported by the collision checker, the relevant cross-unit constraints, and the layout memory summarizing previous rounds. The Critic determines whether the layout contains hard physical errors (deep interpenetration, out-of-bounds placement) or semantic errors (implausible positions or orientations), and outputs a binary modification decision together with qualitative, coordinate-free suggestions (e.g., ``move the chair slightly closer to the table''), as well as an overall layout quality score in $[0, 100]$. Restricting suggestions to qualitative guidance prevents the Critic from anchoring on inaccurate numeric estimates and cleanly separates diagnosis from actuation. Out-of-bounds objects detected geometrically are additionally converted into mandatory correction notes and merged into the Critic's suggestions.

\vspace{-0.7em}

\paragraph{Actor.}
The Actor translates the Critic's suggestions into concrete pose updates. Instead of emitting raw coordinates, the Actor outputs symbolic pose expressions that reference the current layout state. These expressions are resolved by a sandboxed arithmetic evaluator against the current 2D layout, which makes relational adjustments (alignment, symmetric spacing, relative offsets) exact rather than approximated, and confines all edits to interpretable 2D translations and in-plane rotations. Only objects explicitly mentioned in the suggestions are modified; all remaining objects and the anchor are kept fixed.

\vspace{-0.7em}

\paragraph{Layout Memory.}
For every local layout under evolution, we maintain a per-round memory recording the initial pose, the Critic's suggestions, the Actor's pose expressions, the resolved poses, and the detected collisions. The serialized memory is injected into subsequent Critic and Actor prompts. The Critic is explicitly instructed to inspect this memory for repeated failure patterns (e.g., left-right oscillation of the same object along one axis, or repeated boundary-violating moves) and, upon detection, to avoid repeating the failed fix or its symmetric reversal, pivoting instead to a qualitatively different remedy such as adjusting another axis, applying a more decisive move, or targeting a different object. This mechanism is the primary safeguard against the cyclic-adjustment deadlocks described in the main paper.

\vspace{-0.7em}

\paragraph{Convergence and Round Selection.}
Evolution of a local layout terminates when (i) the Critic requests no further modification, (ii) the maximum number of rounds is reached, or (iii) an early-stopping criterion is met: the maximum per-object translation between consecutive rounds falls below a threshold, rotations remain unchanged, and neither collisions nor out-of-bounds objects are detected. After termination, if the final round is free of physical issues it is selected; otherwise, the round with the highest physical quality (fewest collision pairs and out-of-bounds objects, with later rounds preferred at ties) is selected as the output of this variant. The selected layout then undergoes a final gravity-based simulation with the anchor as the ground body to remove any residual floating or penetration before entering the composition stage.

\section{Cross-Unit Constraint Extraction}

As described in the main paper, local units are evolved independently, so sparse cross-unit constraints are required to preserve global coherence. A dedicated agent analyzes the scene image together with the anchor-grouped object relations and emits constraints of two types: \textit{orientation alignment} (e.g., floor-standing chairs must face the desk, while desk-mounted monitors must share the desk's orientation) and \textit{placement alignment} (e.g., a wall-mounted TV and a floor-standing TV stand must be horizontally centered on the same wall). Two design rules keep these constraints reliable. First, a constraint is generated only when the two objects have a strong functional association admitting a unique reasonable configuration; weakly associated or multi-solution pairs (e.g., a painting and a sofa) are deliberately excluded so as not to suppress layout diversity. Second, since each local unit is evolved in isolation, a constraint may only reference the unit's own anchor or objects within the same unit; relations that span two units are decomposed into per-unit clauses through a shared intermediate reference (typically the anchor itself). The resulting constraints are grouped by local unit and injected into the corresponding variant-generation, Critic, and Actor prompts, where the Critic explicitly verifies their satisfaction at every round.

\section{Combinatorial Assembly and Validation}

\paragraph{Local Variant Filtering.}
Before assembly, every evolved local layout (including the refined original) is validated in isolation: we run the collision checker on its post-simulation layout and verify that all target objects remain within the anchor's boundary under orthographic projection. Non-original variants exhibiting residual collisions or out-of-bounds objects are discarded, so that only physically sound local layouts enter the Cartesian combination.

\vspace{-0.7em}

\paragraph{Anchor-Relative Composition.}
Local layouts are expressed in the coordinate frame induced by their own anchor, which allows a variant of a child unit to be transplanted under any variant of its parent unit. Writing the layout parameters $\mathbf{T} = (\mathbf{p}, \mathbf{R}, \mathbf{s})$ of each object as a homogeneous transform $\mathbf{M}$, a child unit whose anchor has local transform $\mathbf{M}_{a}^{\text{local}}$ is grafted onto the parent scene, in which the same anchor has been placed at $\mathbf{M}_{a}^{\text{scene}}$, by applying
\begin{equation}
    \mathbf{M}_i^{\text{scene}} = \mathbf{M}_{a}^{\text{scene}} \left(\mathbf{M}_{a}^{\text{local}}\right)^{-1} \mathbf{M}_i^{\text{local}}
\end{equation}
to every object $o_i$ in the child unit, followed by a polar decomposition to recover valid $(\mathbf{p}_i, \mathbf{R}_i, \mathbf{s}_i)$. Candidate scenes are enumerated by recursively expanding the scene tree from its root anchors and taking the Cartesian product of variant choices across sibling subtrees, yielding the candidate pool $\mathcal{C}$ of $k$ globally complete scenes. Each assembled candidate is subjected to a final scene-level collision check, and candidates with any residual interpenetration are removed from $\mathcal{C}$ before selection.

\vspace{-0.7em}

\begin{algorithm}[!t]
\caption{Representative Scene Selection via Dynamic Max-Min Greedy Search}
\label{alg:scene_selection}
\KwIn{Candidate scene set $\mathcal{C}$, Original baseline scene $S_{\text{org}}$, Target variant count $m$, Filtering factor $\alpha$, Weights $(w_p, w_u, w_r)$}
\KwOut{Selected representative scene set $\mathcal{R} = \{S_{\text{org}}, S^{(1)}, S^{(2)}, \dots, S^{(m)}\}$}

\BlankLine
\tcp{Step 0: Initialize with the baseline original scene}
$\mathcal{R} \leftarrow \emptyset$ \;
\If{$S_{\text{org}} \in \mathcal{C}$}{
    $\mathcal{R} \leftarrow \mathcal{R} \cup \{S_{\text{org}}\}$ \;
    $\mathcal{C} \leftarrow \mathcal{C} \setminus \{S_{\text{org}}\}$ \;
}

\BlankLine
\tcp{Step 1: Iterative Max-Min Greedy Selection with Dynamic Pruning}
\While{$|\mathcal{R}| < m+1$ \textbf{and} $\mathcal{C} \neq \emptyset$}{
    $d_{\max} \leftarrow -\infty$ \;
    $S^* \leftarrow \text{null}$ \;

    \tcp{Find candidate maximizing the minimum distance to selected set}
    \For{\textbf{each} $S \in \mathcal{C}$}{
        $d_{\min}(S) \leftarrow \min_{S_r \in \mathcal{R}} \mathcal{N}(S_r, S)$ \;
        \If{$d_{\min}(S) > d_{\max}$}{
            $d_{\max} \leftarrow d_{\min}(S)$ \;
            $S^* \leftarrow S$ \;
        }
    }

    \If{$S^* == \text{null}$}{
        \textbf{break} \;
    }

    \tcp{Add selected candidate and update pools}
    $\mathcal{R} \leftarrow \mathcal{R} \cup \{S^*\}$ \;
    $\mathcal{C} \leftarrow \mathcal{C} \setminus \{S^*\}$ \;

    \BlankLine
    \tcp{Dynamic per-round pruning}
    $\tau \leftarrow d_{\max} \cdot \alpha$ \;
    $\mathcal{C} \leftarrow \left\{ S \in \mathcal{C} \;\middle|\; \min_{S_r \in \mathcal{R}} \mathcal{N}(S_r, S) \ge \tau \right\}$ \;
}

\Return $\mathcal{R}$ \;
\end{algorithm}

\paragraph{Representative Scene Selection.}
Algorithm~\ref{alg:scene_selection} summarizes the Dynamic Max-Min Greedy Search introduced in the main paper. The pairwise distances are computed with the novelty-aware scene distance using weights $(w_p, w_u, w_r) = (0.4, 0.4, 0.2)$. Since the normalization denominators of $D_{\mathrm{pos}}$ and $D_{\mathrm{dist}}$ are measured with respect to scene $A$, the scene distance is not symmetric, and we adopt a fixed direction convention: whenever a candidate $S \in \mathcal{C}$ is evaluated against a selected scene $S_r \in \mathcal{R}$, the selected scene always serves as the reference $A$, i.e., we compute $\mathcal{N}(S_r, S)$. As the selected set is seeded with the original scene $S_{\text{org}}$, all novelty values are thus uniformly expressed as relative deviations from already-confirmed layouts. All evaluations are cached by ordered pair so that each required distance is measured at most once. We use a filtering factor $\alpha = 0.1$ and select $m = 5$ variant scenes (in addition to the base scene) in all experiments. The dynamic pruning step removes, after each selection, all candidates whose minimum distance to the current selected set $\mathcal{R}$ falls below $\tau = \alpha \cdot d_{\max}$; this both prevents near-duplicates of already selected scenes from being revisited and reduces the number of distance evaluations in later rounds. Since the baseline scene $S_{\text{org}}$ is seeded into $\mathcal{R}$ first, all subsequent selections are implicitly encouraged to differ from the reference layout as well as from each other.

\section{Diversity Evaluation}

Since repeated runs of a baseline yield different reconstructed assets, their variation conflates asset sampling with layout rearrangement, and diversity cannot be compared across methods directly. We therefore evaluate against degraded variants of our own pipeline, all sharing the same assets and producing $m = 5$ layouts per input. Div-IoU averages the pairwise IoU of top-down occupancy maps over the $\binom{5}{2} = 10$ pairs of selected layouts (lower is more diverse), while Div-N applies the novelty-aware distance $\mathcal{N}$ to the same pairs under the direction convention above (higher is more diverse). Div-IoU serves as the primary metric, as Div-N is the objective our greedy selection maximizes. Both are averaged over the 100 scenes of SceneEval-100.

Table~\ref{tab:diversity_pipeline} isolates the components responsible for diversity. Replacing the Dynamic Max-Min Greedy Search with uniform random sampling from $\mathcal{C}$ leaves quality untouched but yields markedly more redundant layouts, confirming that diversity is driven by the selection strategy rather than the candidate pool alone. Repeated sampling is both the least diverse and the most expensive setting, as independent re-sampling tends to rediscover the same locally optimal arrangement.

Table~\ref{tab:diversity_noise} examines whether diversity is simply bought by degrading quality. We perturb the base layout with Gaussian noise on object positions and orientations, followed by the same gravity-based simulation. Diversity increases monotonically with $\sigma$, but semantic quality and structural coherence collapse in step as objects drift off their supporting surfaces or rotate away from their functional partners. At the noise level matched to our own diversity, our method retains substantially higher semantic scores and relation preservation, showing that our diversity is not obtained by trading away layout validity.

\begin{table*}[t]
\centering
\caption{Diversity against degraded variants of our own pipeline. All settings share the same reconstructed assets and produce $m = 5$ layouts per input. \textit{w/o Novelty Selection} draws $m$ scenes uniformly at random from $\mathcal{C}$, and \textit{Repeated Sampling} runs the full agentic evolution $m$ times from the same initialization with different random seeds, without local unit decomposition or Cartesian assembly. Time is amortized per variant.}
\label{tab:diversity_pipeline}
\small
\begin{tabular}{l|cc|c}
\toprule
Method & Div-IoU $\downarrow$ & Div-N $\uparrow$ & Time $\downarrow$ \\
\midrule
Full SceneMosaic & \cellcolor{tabfirst}0.42$\pm$0.08 & \cellcolor{tabfirst}0.37$\pm$0.06 & \cellcolor{tabfirst}0.03h \\
w/o Novelty Selection & 0.58$\pm$0.09 & 0.19$\pm$0.05 & \cellcolor{tabfirst}0.03h \\
Repeated Sampling & 0.71$\pm$0.07 & 0.12$\pm$0.04 & 0.14h \\
\bottomrule
\end{tabular}
\end{table*}

\begin{table*}[t]
\centering
\caption{Diversity versus semantic quality under a random perturbation sweep. RelKeep denotes the fraction of attach/contain relations of the base scene that remain satisfied in the variant. The $\sigma = 0.25$ setting is matched to the diversity of Full SceneMosaic, enabling a like-for-like comparison at equal diversity.}
\label{tab:diversity_noise}
\small
\begin{tabular}{l|cc|ccc}
\toprule
Method & Div-IoU $\downarrow$ & Div-N $\uparrow$ & POS $\uparrow$ & ROT $\uparrow$ & RelKeep $\uparrow$ \\
\midrule
Full SceneMosaic & 0.42$\pm$0.08 & 0.37$\pm$0.06 & \cellcolor{tabfirst}84.5$\pm$2.4 & \cellcolor{tabfirst}79.9$\pm$3.6 & \cellcolor{tabfirst}99.1$\pm$1.2 \\
\midrule
Perturbation, $\sigma = 0.10$ & 0.63$\pm$0.08 & 0.15$\pm$0.05 & 79.2$\pm$3.5 & 71.4$\pm$4.6 & 88.3$\pm$4.2 \\
Perturbation, $\sigma = 0.25$ & 0.43$\pm$0.09 & 0.30$\pm$0.07 & 71.6$\pm$4.4 & 62.1$\pm$5.3 & 71.5$\pm$6.1 \\
Perturbation, $\sigma = 0.50$ & \cellcolor{tabfirst}0.24$\pm$0.10 & \cellcolor{tabfirst}0.52$\pm$0.09 & 60.3$\pm$5.6 & 51.7$\pm$6.2 & 52.8$\pm$7.4 \\
\bottomrule
\end{tabular}
\end{table*}

\section{Limitations}

As SceneMosaic relies on image-based initialization, it inherits an intrinsic constraint of image-to-3D models: a single image can typically depict only one room, restricting generation to room-scale scenes. Consequently, our framework cannot directly produce multi-room houses or entire buildings. Extending the framework to compose room-level generations into coherent house-scale environments is a promising direction.

\section{Prompt Design for Layout Evolution}

We provide the full texts of the role-specific prompts used in our framework. Prompt~\ref{lst:prompt_variant} proposes alternative layout archetypes for each local unit. Prompt~\ref{lst:prompt_pose} instantiates each archetype into concrete 2D poses; for containment units, a variant of this prompt is used in which rotations are frozen and only object centers are predicted. Prompts~\ref{lst:prompt_critic} and \ref{lst:prompt_actor} correspond to the Critic and the Actor in the agentic evolution loop, respectively. Prompt~\ref{lst:prompt_cross} extracts cross-unit constraints prior to evolution. At runtime, each prompt is concatenated with the anchor constraint, the current 2D layout state, and, where applicable, the cross-unit constraints and the layout memory.

\begin{lstlisting}[style=promptstyle, caption={Prompt for alternative layout archetype proposal.}, label={lst:prompt_variant}]
# Alternative Scene Layout Generation Task

You are an intelligent spatial design and architectural planning agent. Your task is to conceptualize and generate alternative, highly distinct layout configurations for a 2D orthogonal scene based on an existing environment layout. 

Instead of just shifting a single object or rotating the entire room, you must generate entirely new, structurally diverse layout archetypes that represent different functional or stylistic design philosophies.

### Input Data:
1. **Current Scene Image**: An orthogonal camera rendering of the current environment with a coordinate grid or reference markers.
2. **Current Layout Data (JSON)**: 2D center coordinates (u, v) and orientations of all existing objects in the scene.
3. **Cross Anchor Constraints (optional)**: Additional layout constraints grouped by anchor_id.

### Objective:
Generate a set of distinctly different, highly representative alternative layouts. Each layout must match a specific spatial design pattern.

### Placement Constraint:
All objects must be realistically supported by the anchor's surface. Ensure accurate bounding box calculations to prevent objects from overlapping or extending beyond the anchor's physical edges.
If cross anchor constraints are provided, the generated alternative layouts should strictly satisfy them.

### Output Requirements:
Return a JSON object containing alternative layout variations. The number of alternative layouts to generate depends on the total number of objects in the current layout:
- **1-5 objects**: Generate 0 to 1 alternative layout
- **6-10 objects**: Generate 0 to 2 alternative layouts
- **More than 10 objects**: Generate 0 to 3 alternative layouts

Do not include coordinate values in the descriptions of the layout variations. **If the existing layout is determined to be the only representative layout, return the alternative layouts array as empty.**

The output must strictly follow this structure (with an example):

```json
{
  "alternative_layouts": [
    {
      "layout_id": 1,
      "layout_style_name": "Centralized Focal Point",
      "layout_reasoning": "The main sofa is positioned flush against the wall, directly facing the TV stand anchored to the opposite side. The coffee table is centered between them, leaving the perimeter walkways completely clear for traffic."
    },
    {
      "layout_id": 2,
      "layout_style_name": "string",
      "layout_reasoning": "brief description"
    }
  ]
}
\end{lstlisting}

\begin{lstlisting}[style=promptstyle, caption={Prompt for 2D pose instantiation.}, label={lst:prompt_pose}]
# Object Placement and Orientation Task

You are an intelligent spatial planning agent. Your task is to determine the optimal placement for a new object within a given 2D orthogonal scene based on the provided visual and textual information.

### Input Data:
1. **Scene Image**: An orthogonal camera rendering of the current layout with a coordinate grid or reference markers. Blue arrows in the rendered image indicate the facing orientation of each object.
2. **Layout Description**: A textual description of where each object is placed in this layout and explanation of why they are placed there.
3. **Current Layout Data (JSON)**: Object centers, orientations, and bounding boxes in grid coordinates.
4. **Cross Anchor Constraints (optional)**: Additional layout constraints grouped by anchor_id.

### Objective:
Based on the new layout description and existing layout examples, generate a layout that is different from the existing one, identifying the precise 2D center coordinates (u, v) and the clockwise rotation angle (in degrees) around the axis perpendicular to the screen for the object.
Coordinate convention: u increases to the right, v increases upward; rotation directions use 0 for facing down, 90 for facing left, 180 for facing up, and -90 for facing right.

### Placement Constraint:
All objects must be realistically supported by the anchor's surface. Ensure accurate bounding box calculations to prevent objects from overlapping or extending beyond the anchor's physical edges.
The rotation angle represents the object's facing direction, so it must follow common sense and the scene context: for example, chairs should face toward the center of the table they are placed around, and objects near the edge of a floor should face inward toward the interior of the floor.
If cross anchor constraints are provided, the generated pose should strictly satisfy them.

### Output Requirements:
Return a JSON object with the following structure:
```json
{
  "poses": [
    {
      "object_id": "string",
      "reasoning": "Placed based on text description. Coordinates and rotation are calculated using bounding boxes to fit within the anchor's physical edges and avoid overlapping.",
      "center_coordinates": {
        "u": float,
        "v": float
      },
      "rotation_degrees": float
    },
    {
      "object_id": "string",
      ...
    }
  ]
}

Notes:
- The rotation angle is absolute. Set this value to its original angle if no orientation change is required.
- Do not change the anchor object.
- Use precise numeric values for `u`, `v`, and `rotation_degrees`.
- When generating layouts, explicitly compute the boundaries of each object to avoid collisions.
\end{lstlisting}

\begin{lstlisting}[style=promptstyle, caption={Prompt for the Critic (layout evaluation).}, label={lst:prompt_critic}]
# Layout Evaluation Task

You are an intelligent spatial evaluation agent. Your task is to judge whether a rendered layout has any hard errors in physical collisions or semantic (unreasonable) placement based on the rendered image and current layout. If there are any errors, please provide suggestions for corrections.

### Input Data:
1. **Rendered Scene Image**: An orthogonal camera rendering of the current layout with grid overlay. Blue arrows in the rendered image indicate the facing orientation of each object.
2. **Current Layout Data (JSON)**: Object centers, orientations, and bounding boxes in grid coordinates.
3. **Potential Deep Collisions (optional)**: Pairs flagged by physics contact checks with contact counts.
4. **Layout Memory (optional)**: Previous rounds of modification notes, rejected suggestions, revised poses, and collision history for this same layout group.
5. **Cross Anchor Constraints (optional)**: Additional layout constraints grouped by anchor_id.

### Objective:
1. Decide whether the current layout has any hard physical collision errors or semantic (unreasonable) placement errors. If a modification is needed, provide high-level guidance without any exact numeric coordinates. Use the direction guidance with the coordinate convention: u increases to the right, v increases upward; rotation directions use 0 for facing down, 90 for facing left, 180 for facing up, and -90 for facing right.
2. Also provide an overall layout quality score in `quality_score` as an integer between 0 and 100 (higher is better). The score should capture: physical collisions severity, semantic placement reasonableness, spacing and accessibility, adherence to anchor constraints, and surface-boundary compliance. When computing the score, weigh hard physical collisions and semantic errors highest (they should strongly reduce the score); minor spacing or alignment issues reduce the score modestly.
3. Explicitly inspect both object rotation degrees and the blue arrows in the rendered image. For floor anchors in particular, check whether edge objects are facing inward. Also re-check whether any rotation-related constraints mentioned in `cross anchor constraints` are being satisfied.

### Constraints:
- Do not change the anchor object.
- Avoid precise numeric values; use qualitative directions only (e.g., "slightly left", "closer to the wall", "facing down").
- Keep objects within the anchor surface boundaries and avoid overlaps.
- If cross anchor constraints are provided, check whether they are satisfied. If any constraint is not satisfied, include it in `modification_notes`.
- Use one of the following two options, not both:
  - If modifications are needed for a single object, you may freely change its position and optionally its rotation.
  - If modifications are needed for a type of object, you may rotate that type in place or apply one uniform translation to all instances of that type, but you must not specify different movements per object; for example, make all chairs face the table, or move all chairs slightly upward by the same amount.
- When you mention any object in a suggestion, include its object ID in parentheses.
- Inspect `LAYOUT_MEMORY` before proposing any correction: If the memory reveals a repeated failure pattern (for example, left-right oscillation on one axis, or repeated boundary-violating moves) for the same object or object pair, explicitly state this reasoning and avoid repeating the failed fix or making a symmetric reversal; instead, pivot to a qualitatively different remedy, such as adjusting a different axis, applying a larger or more decisive move, targeting a different object, or making no modification if no safe alternative exists.

### Output Requirements:
Return a JSON object with the following structure:
```json
{
  "should_modify": true or false,
  "reasoning": "Explanation of why changes are needed.",
  "modification_notes": [
    "Move the chair (object_000) slightly closer to the table (object_001)."
  ],
  "quality_basis": "Explanation of basis for the score.",
  "quality_score": 0-100
}
```

If no changes are needed, set `should_modify` to `false` and return an empty `modification_notes` list.
\end{lstlisting}

\begin{lstlisting}[style=promptstyle, caption={Prompt for the Actor (layout modification).}, label={lst:prompt_actor}]
# Layout Modification Task

You are an intelligent spatial planning agent. Your task is to apply modification guidance to update object poses based on the rendered image, layout description, current layout data, and modification notes.

### Input Data:
1. **Rendered Scene Image**: An orthogonal camera rendering of the current layout with a coordinate grid. Blue arrows in the rendered image indicate the facing orientation of each object.
2. **Current Layout Data (JSON)**: Object centers, orientations, and bounding boxes in grid coordinates.
3. **Modification Guidance**: Qualitative suggestions for how to adjust objects.

### Objective:
Generate updated 2D center coordinates and rotation angles for objects that need adjustment. Use symbolic expressions and raw numeric values flexibly, referencing existing 2D pose values (center, size_2d, bbox_2d) from CURRENT_LAYOUT_DATA when appropriate. Keep objects within the anchor surface. Keep the anchor object unchanged.

### Output Requirements:
Return a JSON object with reasoning and symbolic pose updates. Only include objects that should change.
Example with comments (actual output should be valid JSON without comments):
```jsonc
{
  "reasoning": "Short explanation tied to the modification guidance.",
  "pose_expressions": [
    "object_001.center.u = object_000.center.u", // Align object_001 with object_000 along the u axis
    "object_002.center.u = object_002.center.u + 0.1", // Move object_002 slightly to the right
    "object_003.center.u = 0.2" // Set object_003's u directly to 0.2
    "object_004.rotation_degrees = 90" // Orient object_004 to the left
  ]
}
```

Notes:
- Coordinate convention: u increases to the right, v increases upward; rotation directions use 0 for facing down, 90 for facing left, 180 for facing up, and -90 for facing right.
- Do not change the anchor object.
- Only modify objects explicitly mentioned in the modification guidance; keep all other objects unchanged.
- Use expressions or raw numeric values as appropriate.
- Allowed references: `object_id.center.u`, `object_id.center.v`, `object_id.size.u`, `object_id.size.v`,
  `object_id.bbox.min_u`, `object_id.bbox.min_v`, `object_id.bbox.max_u`, `object_id.bbox.max_v`,
  `object_id.rotation_degrees`. Aliases: `center_2d`, `size_2d`, `bbox_2d`.
- Please explicitly compute object boundary extents to avoid overlap when adjusting poses.
\end{lstlisting}

\begin{lstlisting}[style=promptstyle, caption={Prompt for cross-unit constraint extraction.}, label={lst:prompt_cross}]
# Cross-Anchor Spatial Constraint Generation Task

You are an intelligent spatial harmony agent specializing in defining the rules and constraints necessary for establishing global coherence in a combined scene layout. Your task is to generate **critical cross-anchor constraints** for objects belonging to different anchors in a given scene template or description, ensuring they align or orient correctly when combined.

### What is an Anchor:
Each **anchor** represents a physical surface or container in the scene where objects can be placed. Each anchor defines a localized environment containing a specific set of target objects, within which the model remains completely oblivious to any external objects.

### Context:
Scene layouts are often generated independently for each anchor and then programmatically combined. To prevent global disharmony when these anchors merge, you must generate strict relationship rules (constraints) between objects residing on **different anchors**.

### Input Data:
You will be provided with three key inputs to analyze the environment:
1. **Anchor-Based Relations**: JSON data detailing existing parent-child placement dependencies.
2. **Rendered Visual Map**: A layout visualization showing the spatial arrangement of objects.

### Objective:
Generate a list of critical cross-anchor constraints. Focus exclusively on two primary relationship types:

1. **Orientation Alignment**: Defining how objects on different anchors must face each other (e.g., chairs on the floor must face the opposite direction of a desk; monitors on a desk must face the same way as the desk).
2. **Spatial/Semantic Placement Alignment**: Defining positional or proximity dependencies between different anchors (e.g., a whiteboard on a wall anchor must align centrally with a podium anchor on the floor).

### Constraint Generation Guidelines:
- **Strong Association Only**: Only define a constraint if the two cross-anchor objects have a direct, functional, or strong semantic relationship. Crucially, an alignment constraint should only be included if it represents the single, uniquely reasonable layout configuration. To better illustrate this principle, consider the following specific examples:
  * *Exclude (Weak associations):* A painting/panel on a wall and a sofa on the floor.
  * *Exclude (Multi-solution/Ambiguous scenarios):* A wall-mounted display screen and a conference table in the middle of a meeting room (where multiple positioning layouts are acceptable).
- **Valid References**: Your `layout_note` instructions can *only* use the anchor itself or target objects belonging to its *own* anchor as reference points for positioning and orientation. 
  *(Example: Instead of saying "chairs should face the monitors on the desk", you must state "chairs should face the opposite direction of the desk", because the monitors are invisible when generating the floor anchor.)*
- **Phrasing and Language Standards**: Keep instructions simple, direct, and avoid overly complex commands. Use the following standardized phrasing patterns:
  * *For Orientation/Rotation:* Express clearly using active commands to direct facing angles (e.g., "make Object A face the same/opposite direction as Object B", or "orient Object A to Object B").
  * *For Positioning:* Use explicit spatial relations, such as "keeping Object A stationary", or "positioning Object A at the center / to the left of Object B".
- **Related Object Formatting**: Like the following examples, when mentioning `related_objects` in `layout_note`, place each object id immediately after its category name.
- **Distinct Anchors Only**: For each output relationship constraint, `anchor_id` in `anchor_a` and `anchor_id` in `anchor_b` **must not** be the same.

### Output Requirements:
Return a JSON array containing the identified cross-anchor constraints using the exact structure below:

```json
[
  {
    "description": "The orientation of the chairs on the floor and the monitors on the desk need to face each other. Therefore, using the desk as the intermediate object, the chairs on the floor should face the desk, and the monitors on the desk should share the same orientation as the desk.",
    "anchor_a": {
      "anchor_id": "floor",
      "related_objects": ["object_001", "object_007"],
      "layout_note": "The desk (object_001) and chair (object_007) must always face each other."
    },
    "anchor_b": {
      "anchor_id": "object_001",
      "related_objects": ["object_002", "object_003"],
      "layout_note": "The monitors (object_002, object_003) must always maintain the same facing direction as the desk."
    }
  },
  {
    "description": "The Wall TV and the TV stand on the floor must align. Since the TV is on the wall anchor and the TV stand is on the floor anchor, they use the physical wall surface as an intermediate reference. Both the TV stand and the Wall TV must be centered horizontally on the same wall.",
    "anchor_a": {
      "anchor_id": "floor",
      "related_objects": ["object_010"], 
      "layout_note": "Place the TV stand (object_010) flat against the floor-to-wall intersection edge, centered horizontally along its length."
    },
    "anchor_b": {
      "anchor_id": "wall_001",
      "related_objects": ["object_011"],
      "layout_note": "Mount the TV (object_011) on wall_001, centered horizontally relative to the wall surface."
    }
  }
]
\end{lstlisting}

\end{document}